\documentclass[10pt,twocolumn,letterpaper]{article}

\usepackage[letterpaper,
            textheight=8.875in,
            textwidth=6.875in,
            columnsep=0.3125in,
            top=1in,
            headheight=0pt,
            headsep=0pt]{geometry}

\usepackage{times}

\usepackage{graphicx}
\usepackage{amsmath}
\usepackage{amssymb}
\usepackage{booktabs}
\usepackage[format=plain,labelformat=simple,labelsep=period,font=small,compatibility=false]{caption}
\usepackage[font=footnotesize,skip=3pt,subrefformat=parens]{subcaption}
\usepackage{cite}

\usepackage[hyphens]{url}
\usepackage[breaklinks,colorlinks,citecolor=blue,linkcolor=blue,urlcolor=blue]{hyperref}

\usepackage[capitalize]{cleveref}
\crefname{section}{Sec.}{Secs.}
\Crefname{section}{Section}{Sections}
\Crefname{table}{Table}{Tables}
\crefname{table}{Tab.}{Tabs.}

\makeatletter
\renewcommand{\@maketitle}{%
  \newpage
  \null
  \vskip .375in
  \begin{center}
    {\Large \bf \@title \par}
    \vspace*{24pt}
    {\large
      \lineskip .5em
      \begin{tabular}[t]{c}
        \@author
      \end{tabular}\par}
    \vskip .5em
    \vspace*{12pt}
  \end{center}
}

\font\elvbf  = ptmb scaled 1100
\def\cvprsection{\@startsection {section}{1}{\z@}%
   {10pt plus 2pt minus 2pt}{7pt} {\large\bf}}
\def\cvprssect#1{\cvprsection*{#1}}
\def\cvprsect#1{\cvprsection{\hskip -1em.~#1}}
\def\section{\@ifstar\cvprssect\cvprsect}

\def\cvprsubsection{\@startsection {subsection}{2}{\z@}%
   {8pt plus 2pt minus 2pt}{6pt} {\elvbf}}
\def\cvprssubsect#1{\cvprsubsection*{#1}}
\def\cvprsubsect#1{\cvprsubsection{\hskip -1em.~#1}}
\def\subsection{\@ifstar\cvprssubsect\cvprsubsect}
\makeatother

\renewenvironment{abstract}
  {\centerline{\large\bf Abstract}\vspace*{12pt}\itshape}
  {\vspace*{12pt}}

\begin{document}

\title{Deep Clustering for Climate: Analyzing Teleconnections through Learned Categorical States}

\author{Lívia Meinhardt\\
FGV EMAp\\
{\tt\small livia.meinhardt@fgv.edu.br}
\and
Dário Oliveira\\
FGV EMAp\\
{\tt\small dario.oliveira@fgv.br}
}
\maketitle
\thispagestyle{empty}

\begin{abstract}
Understanding and representing complex climate variability is essential for both scientific analysis and predictive modeling. However, identifying meaningful climate regimes from raw variables is challenging, as they exhibit high noise and nonlinear dependencies. In this work, we explore the use of Masked Siamese Networks to discretize climate time series into semantically rich clusters. Focusing on daily minimum and maximum temperature, we show that the resulting representations: (i) yield clusters that reflect meaningful climate states under our modeling assumptions, offering a simplified representation for downstream use; (ii) enable sampling and analysis of specific climate scenarios; and (iii) exhibit statistical associations with El Niño events, underscoring their scientific relevance. Our findings highlight the potential of self-supervised discretization as a tool for climate data analysis and open avenues for incorporating richer climate indicators in future work.
\end{abstract}


\section{Methodology}
\label{sec:methodology}

We formulate the discretization task as a joint representation learning and clustering problem. We train a ViT encoder using  the Masked Siamese Network (MSN) \cite{msn}, a self-supervised framework that learns semantic representations by matching unmasked and masked views of the same input. MSN consists of an anchor encoder $f_\theta$ (with trainable weights $\theta$) and a target encoder $f_\xi$ (whose weights $\xi$ are an exponential moving average of $\theta$). The training procedure for a given daily climate sample  $x \in \mathbb{R}^{H \times W \times V}$ proceeds as follows:

\textbf{Multi-View Generation}: We generate multiple views of the input data using Random Resized Cropping as the sole augmentation strategy. This preserves the underlying physical values while varying the spatial scale and coverage. The target view ($x^+$) represents a large crop covering most of the region, while $M \geq 1$ anchor views ($x_i$) correspond to smaller crops where random patches (15\% of the grid) are masked.

\textbf{Representation and Prototype Assignment}: The encoders map the views to latent representations $z = f_\theta(\hat{x})$ and $z^+ = f_\xi(x^+)$. We discretize the latent space using $K$ learnable prototypes ${q_k}_{k=1}^K$. The probability of assigning a representation $z$ to prototype $k$ is computed via a softmax with temperature $\tau$:

$$p_k(z, \tau) = \frac{\exp(\langle z, q_k \rangle / \tau)}{\sum_{j=1}^K \exp(\langle z, q_j \rangle / \tau)}.$$

The target assignment $p^+$ uses a low temperature $\tau^+$ to generate sharp pseudo-labels, while the anchor prediction $p$ uses a higher temperature $\tau$.

\textbf{Training Objective}: The model minimizes the cross-entropy between the target assignment and the anchor prediction. In addition, it incorporates Mean Entropy Maximization (ME-MAX) regularization on the average target prediction to ensure utilization of all $K$ clusters and prevent collapse.

\textbf{Inference}: At inference time, the anchor encoder is discarded. We employ the stable target encoder $f_\xi$ to map a full, unmasked input $x$ to a discrete regime. The input is assigned to the prototype $q_k$ that maximizes the cosine similarity with the latent representation $z = f_\xi(x)$. This transformation converts the continuous time series into a deterministic categorical sequence $S = (k^*_1, k^*_2, \dots, k^*_T)$, enabling downstream probabilistic analysis of regime transitions and teleconnections. We refer the reader to \cite{msn} for additional details of Masked Siamease Networks.

\section{Experiments}
\label{sec:experiments}
We evaluate the proposed method on its ability to discover physically meaningful climate regimes and capture large-scale teleconnections without explicit supervision.

\subsection{Dataset}
\begin{figure}
  \centering
   \includegraphics[width=0.8\linewidth]{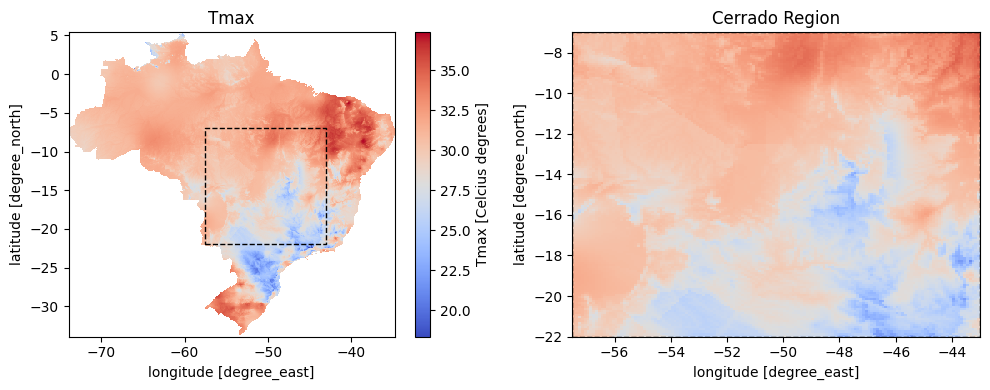}
   \caption{Spatial extent of the selected study region (Brazilian Cerrado) used for model training and evaluation.}
   \label{fig:studyregion}
\end{figure}

We use the Brazilian Daily Weather Gridded Data (BR-DWGD)~\cite{brdwgd} at 0.1° spatial resolution and extract daily minimum (Tmin) and maximum (Tmax) temperature. The dataset is spatially subset to the Brazilian Cerrado using the bounding box latitude $[-22.0, -7.0]$ and longitude $[-57.5, -43.0]$ (as indicated in  Figure~\ref{fig:studyregion}),
resulting in daily samples represented as tensors $x_t \in \mathbb{R}^{H' \times W' \times 2}$. The temporal window spans January 1961 to March 2024. The years 1981 and 2000 are used as test set and all remaining years form the training set.

\subsection{Experimental Design}
We employ a ViT-Small backbone (patch size 16) modified for 2-channel input. We follow the standard training protocol proposed in \cite{msn}, training for 300 epochs using the AdamW optimizer with a cosine learning rate schedule. The  batch size is 512. The projection head maps the encoder output to a 128-dimensional latent space, discretized into $K=30$ prototypes. All experiments were carried out using A100 GPU.

\section{Results}
\subsection{Interpretable Climate Regimes}

\begin{figure*}[t]
    \centering
    \begin{subfigure}{0.45\textwidth}
        \includegraphics[width=\linewidth]{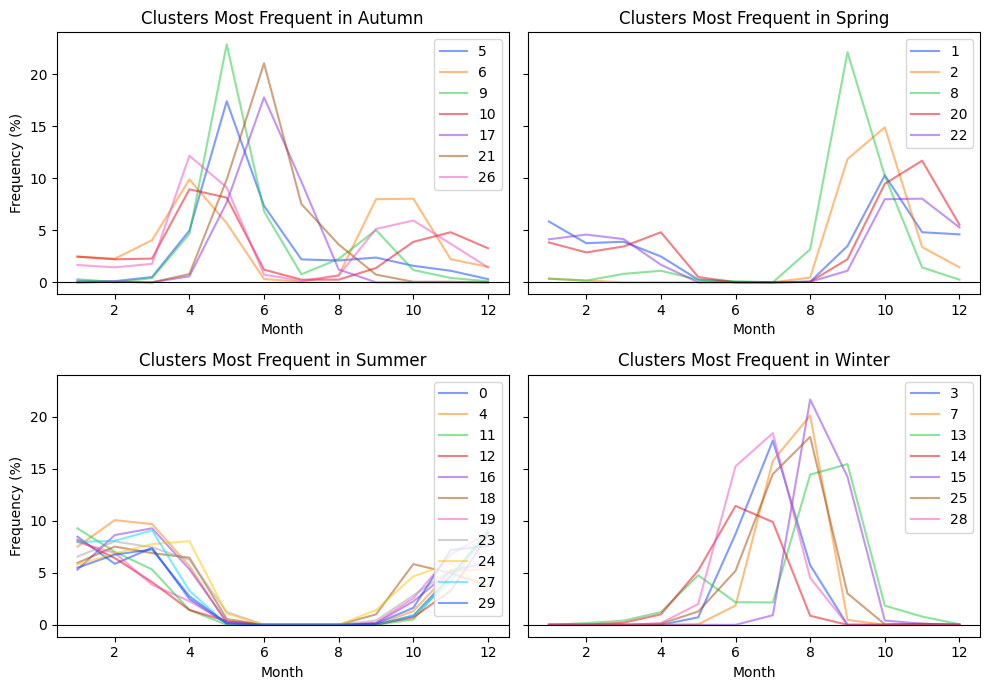}
        \caption{Monthly frequency}
        \label{fig:cluster_frequencies}
    \end{subfigure}\hfill
    \begin{subfigure}{0.45\textwidth}
        \includegraphics[width=\linewidth]{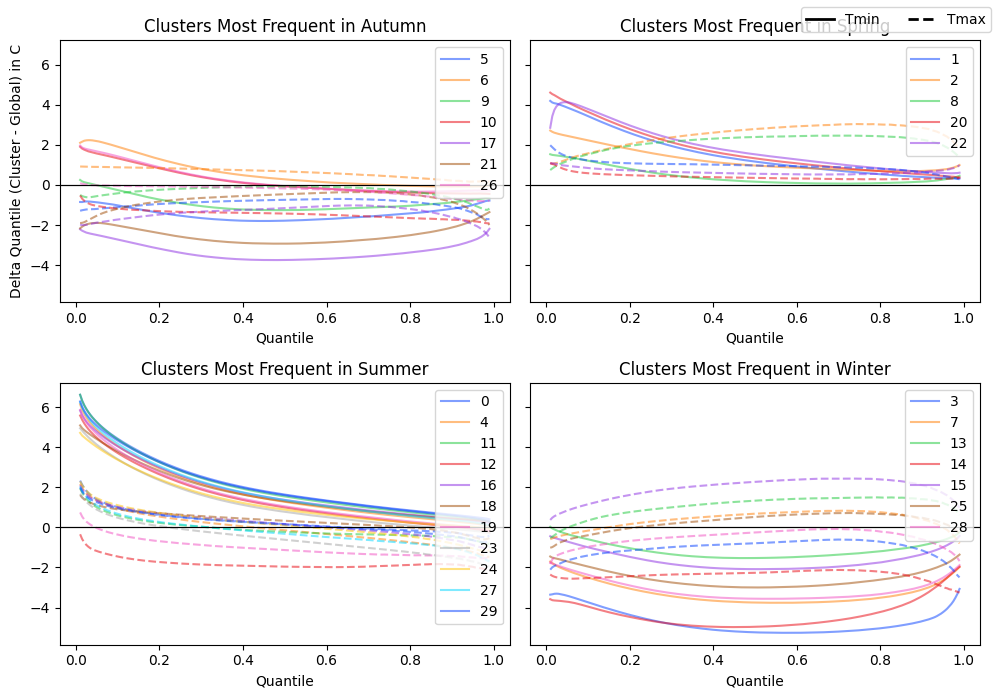}
        \caption{Quantile anomalies ($\Delta$)}
        \label{fig:cluster_quantiles}
    \end{subfigure}
    \caption{Temporal and atmospheric characterization of the learned climate regimes. (a) Monthly frequency of the $K=30$ clusters (\%), grouped into seasonal meta-clusters according to their peak occurrence. (b) Delta-quantile temperature anomalies relative to the full dataset, shown for daily minimum ($T_{\min}$, solid) and maximum ($T_{\max}$, dashed) temperatures. Positive (negative) values indicate warmer (cooler) conditions at a given quantile, in $^\circ$C.}
    \label{fig:clusters_temporal_thermo}
\end{figure*}

The learned prototypes organize the daily temperature fields into a compact set of regimes that are both seasonally structured and thermally distinct. Figure~\ref{fig:clusters_temporal_thermo} summarizes these two aspects. Monthly frequencies reveal a clear annual organization: most of the $K=30$ regimes are concentrated in a single season, while a smaller subset remains more broadly distributed but still exhibits a dominant seasonal preference. Grouping regimes by their month of maximum occurrence yields four seasonal meta-clusters that provide a compact description of the categorical sequence $S$.

Thermal characterization complements this temporal view. The delta-quantile anomalies in Figure~\ref{fig:clusters_temporal_thermo}(b) compare the temperature distribution of each regime against the full dataset for both daily minimum and maximum temperatures. Rather than differing only in their mean state, the regimes separate distinct parts of the temperature distribution, including warm and cold extremes as well as transitional configurations. Together, these results indicate that the self-supervised discretization captures climate states that are not only temporally coherent, but also physically meaningful.

\subsection{Teleconnection Analysis}

ENSO exerts a strong and structured influence on the occurrence of the learned regimes. We define ENSO states from the Oceanic Niño Index (ONI) and examine this influence from three complementary perspectives: aggregate probability anomalies, lagged responses, and month-conditioned behavior. Across these analyses, the response is concentrated in a small subset of regimes and becomes stronger in recent decades.

\paragraph{ENSO-induced changes in regime frequency.}

\begin{figure}[t]
    \centering
    \includegraphics[width=\linewidth]{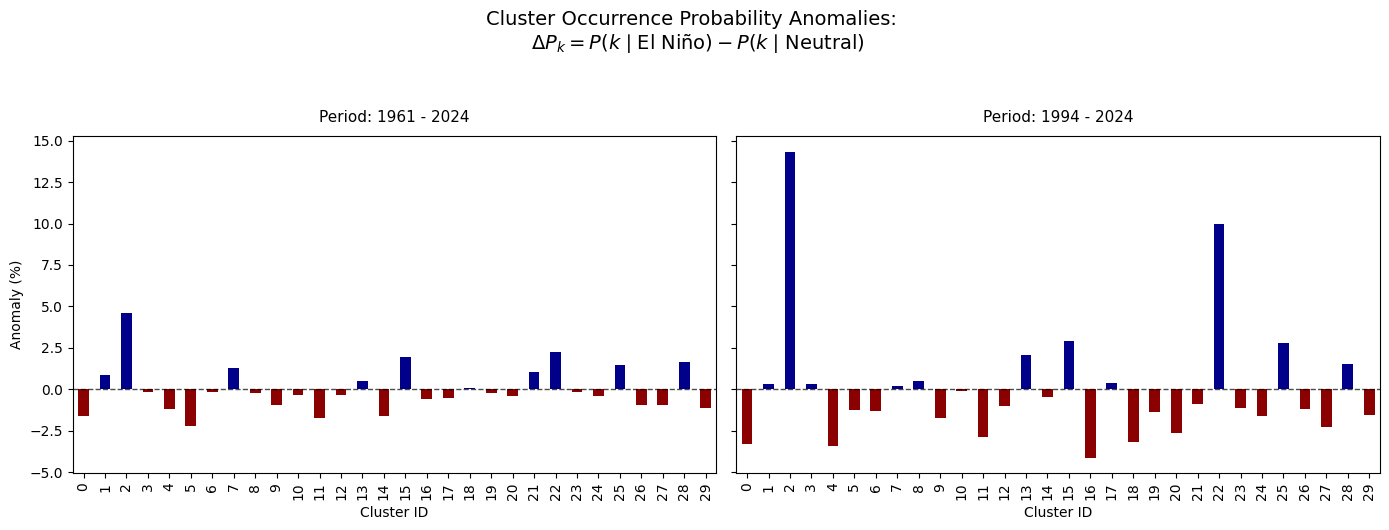}
    \caption{Cluster occurrence probability anomalies ($\Delta P_k$) between El Niño and Neutral conditions for two time periods: the full record (1961--2024) and a recent 30-year subset (1994--2024). Positive values indicate increased regime frequency during El Niño, while negative values indicate suppression.}
    \label{fig:enso_global}
\end{figure}

Figure~\ref{fig:enso_global} shows the anomaly in occurrence probability between El Niño and Neutral conditions, defined as
\[
\Delta P_k = P(k \mid \text{El Niño}) - P(k \mid \text{Neutral}).
\]
The response is dominated by clusters 2 and 22, with substantially weaker anomalies for the remaining regimes. In the full record, both clusters already stand out as the main ENSO-sensitive regimes. Over the most recent 30 years, this contrast becomes much sharper: cluster 2 reaches the largest anomaly, while cluster 22 becomes the second strongest response. This intensification indicates that ENSO teleconnections in the Cerrado are not only detectable in the learned representation, but also non-stationary over time.

These two dominant regimes are both associated with warm conditions, but they differ in their thermal structure. Cluster 2 is linked to strong positive anomalies in the upper tail of maximum temperature, consistent with intense hot extremes. Cluster 22 is more strongly associated with elevated minimum temperature, reflecting a broader and more persistent warming pattern. The model therefore distinguishes different types of ENSO-related heat regimes rather than collapsing them into a single warm state.

\paragraph{Lagged structure of ENSO teleconnections.}

To examine temporal structure, we compute lagged anomalies relative to ONI,
\[
\Delta P_k(\tau) = P(k \mid \text{El Niño}_\tau) - P(k \mid \text{Neutral}_\tau), \tau \in [-12,12].
\]
This analysis reveals that ENSO teleconnections are not instantaneous. Different regimes peak at different lead--lag times, indicating a temporally asymmetric response throughout the ENSO lifecycle. Cluster 22 shows a predominantly synchronous response, with its largest anomaly near zero lag, whereas cluster 2 also exhibits enhanced sensitivity close to the ENSO peak but contributes strongly to the early development phase. Other regimes show precursor or delayed behavior, suggesting that the learned categorical states capture both concurrent and lagged signatures of large-scale forcing.

More broadly, the lagged patterns reveal coordinated groups of regimes whose probabilities rise and fall in opposite phases. This indicates that ENSO modulates the regional climate not by affecting one isolated regime, but by redistributing probability mass across several learned states over time.

This grouped behavior is particularly informative when regimes are interpreted jointly rather than one by one. Winter-associated and summer-associated regimes tend to organize into opposite lag phases, suggesting that ENSO shifts the balance between contrasting regional temperature configurations as the event develops and decays. In this sense, the lagged anomalies describe a structured temporal reallocation of regime probability, not just isolated sensitivity peaks. A grouped visualization of these responses is provided in the appendix (Figure~\ref{fig:anomalies_lags_groups_appendix}), where regimes with similar lag profiles are shown together to highlight coherent oscillatory behavior across the ENSO lifecycle.

\begin{figure}[t]
    \centering
    \includegraphics[width=\linewidth]{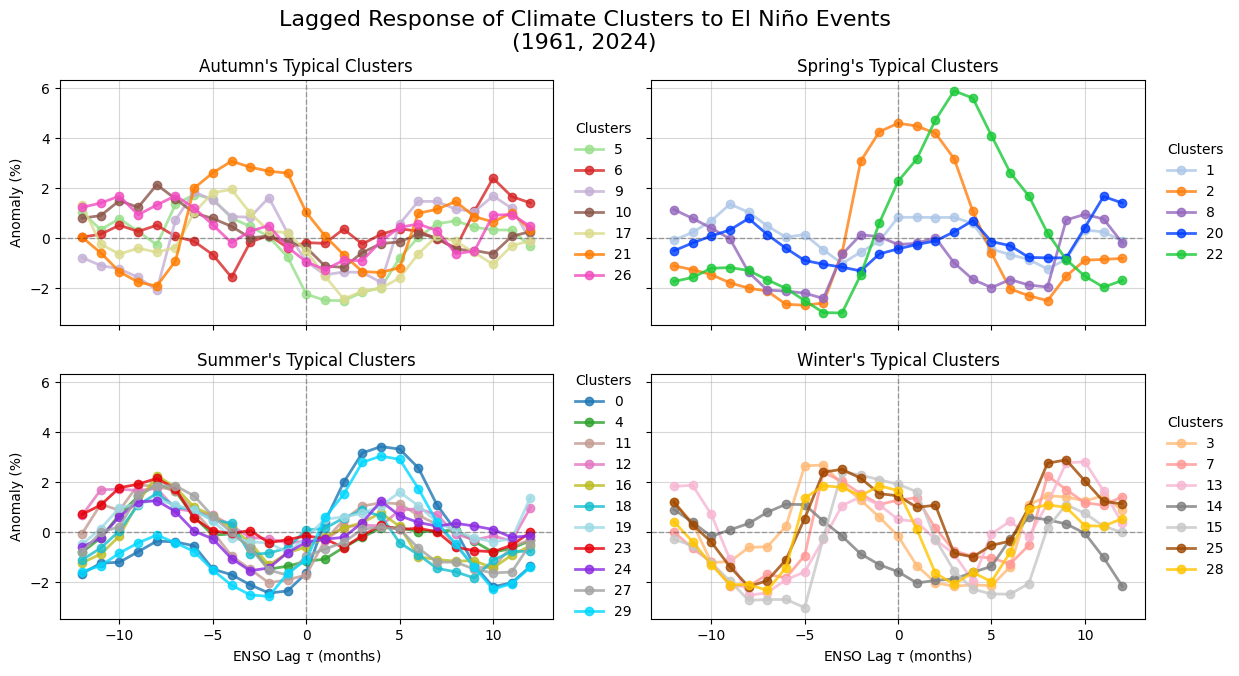}
    \caption{Lagged probability analysis showing the sensitivity of the learned regimes at lags $\tau \in [-12,+12]$ months relative to the Oceanic Niño Index. Negative lags indicate precursor behavior before El Niño, while positive lags indicate delayed responses after the ENSO peak.}
    \label{fig:enso_lags}
\end{figure}

\paragraph{Seasonally controlled ENSO response.}

\begin{figure}[!t]
    \centering
    \begin{subfigure}{\linewidth}
        \centering
        \includegraphics[width=0.96\linewidth]{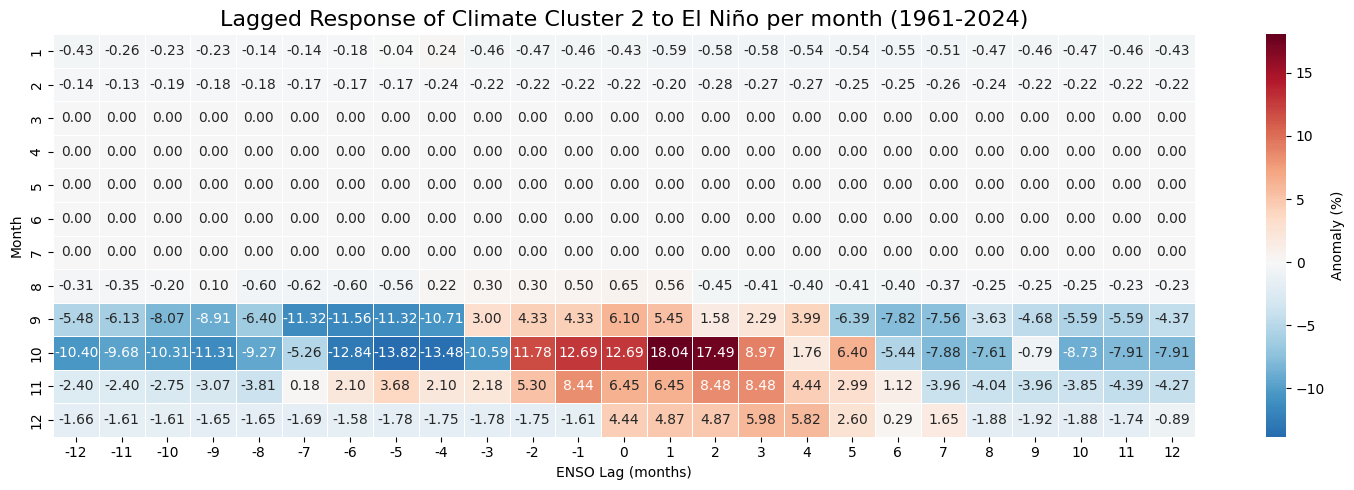}
        \caption{Cluster 2}
    \end{subfigure}

    \vspace{0.3cm}

    \begin{subfigure}{\linewidth}
        \centering
        \includegraphics[width=0.96\linewidth]{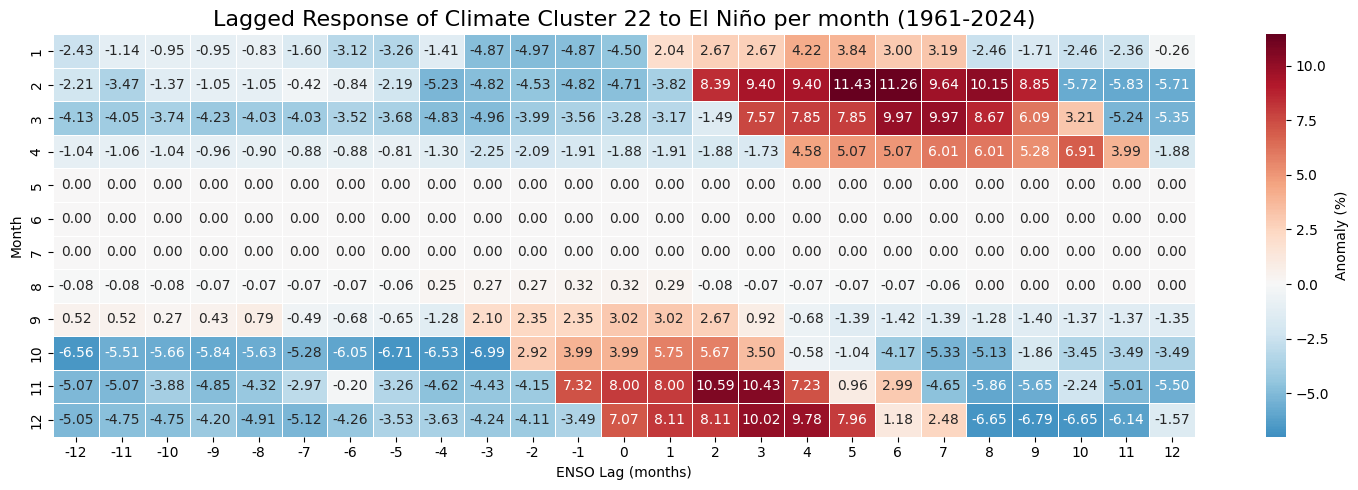}
        \caption{Cluster 22}
    \end{subfigure}
    \caption{Month-conditioned lagged ENSO anomalies for the two dominant ENSO-sensitive regimes. Values represent $\Delta P_k(m,\tau)$, showing how the response varies jointly with calendar month and ENSO lag.}
    \label{fig:enso_monthly}
\end{figure}

Because regime frequencies are strongly seasonal, we next condition the ENSO anomalies on calendar month. For a given month $m$, we compute
\[
\Delta P_k(m,\tau) = P(k \mid \text{El Niño}_\tau, m) - P(k \mid \text{Neutral}_\tau, m).
\]
Figure~\ref{fig:enso_monthly} shows the month-specific lagged response for clusters 2 and 22. The aggregate ENSO signal persists after controlling for seasonality, but its amplitude and timing depend strongly on the background month. Some months show strong amplification, whereas others show suppression, demonstrating that ENSO modifies the annual cycle of regime occurrence rather than simply increasing warm-regime frequency uniformly throughout the year.

This seasonal conditioning clarifies that the dominant ENSO-sensitive regimes are tied to specific phases of the regional climate cycle. In other words, the learned teleconnections are both phase-locked to ENSO and filtered by the local seasonal state.

\paragraph{Long-term evolution of ENSO-sensitive regimes.}

\begin{figure}[ht]
    \centering
    \begin{subfigure}{\linewidth}
        \centering
        \includegraphics[width=0.96\linewidth]{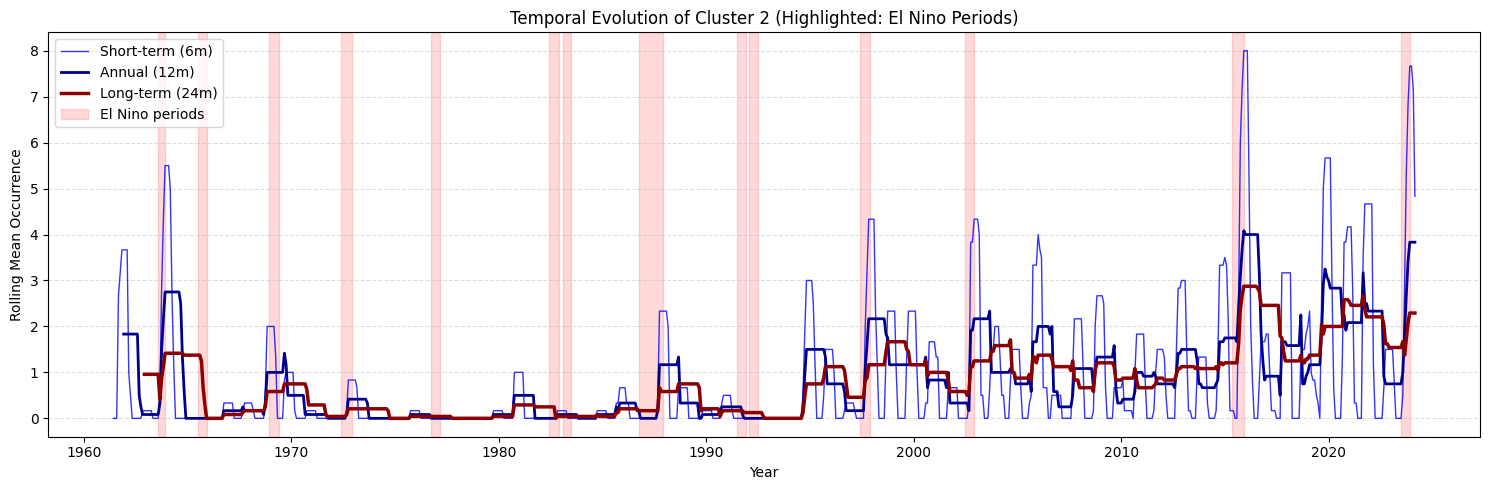}
        \caption{Cluster 2}
    \end{subfigure}

    \vspace{0.3cm}

    \begin{subfigure}{\linewidth}
        \centering
        \includegraphics[width=0.96\linewidth]{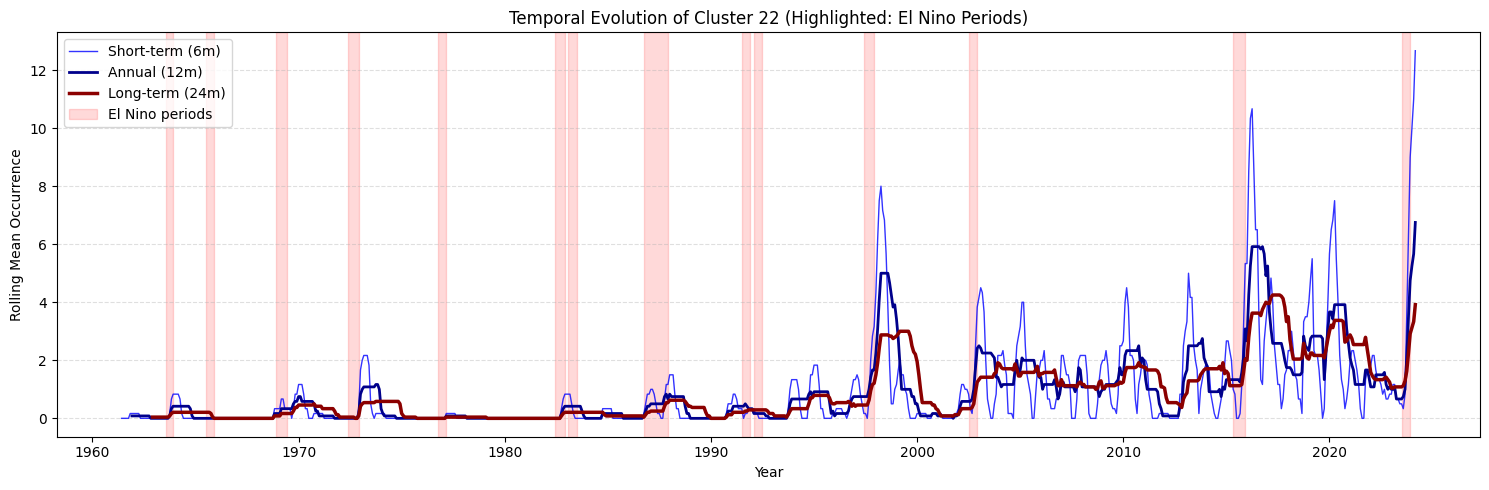}
        \caption{Cluster 22}
    \end{subfigure}
    \caption{Time series of monthly regime frequencies for the two dominant ENSO-sensitive clusters, shown with running means. El Niño periods are indicated for reference.}
    \label{fig:clusters_timeseries}
\end{figure}

Finally, we examine how the frequency of the two dominant ENSO-sensitive regimes evolves over time. Figure~\ref{fig:clusters_timeseries} shows monthly frequencies together with running means for clusters 2 and 22. Both regimes transition from relatively weak and infrequent occurrence early in the record to larger peaks and higher baseline occurrence in recent decades. The increase is especially pronounced after the late 1990s, when both regimes become more persistent and more active during El Niño periods.

Cluster 22 is particularly notable because it is nearly absent during much of the earlier record and only emerges as a recurrent regime in recent decades. Taken together, the stronger recent peaks and higher average occurrence indicate that the ENSO-related extreme regimes have become more prominent over time. This suggests that the learned representation captures a non-stationary teleconnection signal, in which extreme warm states are appearing more often and with greater persistence.

\section{Discussion and Conclusion}

This work shows that self-supervised discretization can recover climate regimes that are both interpretable and useful for teleconnection analysis. The learned states are seasonally organized, thermally distinct, and sensitive to ENSO in ways that are physically plausible. Rather than treating each daily field independently, the framework converts high-dimensional temperature maps into a compact categorical sequence, making it easier to track how regional climate behavior shifts across ENSO phases, lags, and seasons.

The teleconnection analysis reinforces the main value of the proposed framework: it provides a useful and interpretable tool for characterizing regional climate regimes and how they evolve under large-scale forcing. The learned representation highlights both regime-specific and grouped responses to ENSO. At the regime level, clusters 2 and 22 emerge as the dominant warm-extreme states, each capturing a distinct thermal signature. At the grouped level, the analysis shows that ENSO reorganizes coherent sets of winter, summer, autumn, and spring regimes whose probabilities rise and fall in opposite phases across the ENSO lifecycle. Together with the lagged, month-conditioned, and long-term analyses, these results show that the framework captures not only which regimes are ENSO-sensitive, but also how their timing, seasonal modulation, and persistence structure the regional climate response.

More broadly, the results illustrate the value of representation learning for climate science. A discrete regime description offers a practical middle ground between raw gridded fields and highly aggregated scalar indices, preserving enough structure to support interpretation while simplifying downstream analysis. This makes the approach promising for problems where understanding coordinated patterns and their drivers is more important than predicting a single local variable.

Future work should extend the framework beyond temperature alone. Incorporating variables such as precipitation, humidity, or circulation fields could provide a more complete characterization of climate regimes and help identify multi-variable teleconnection signatures. Another important direction is to apply the same methodology to other regions and climate modes, including phenomena such as the Madden--Julian Oscillation or Atlantic variability. Finally, linking learned climate regimes to impact-focused applications, such as agriculture, fire risk, or vector-borne disease dynamics, could help translate these representations into tools for climate risk assessment and environmental decision-making.

\clearpage
\onecolumn
\appendix

\section{Additional Teleconnection Figures}

Figure~\ref{fig:anomalies_lags_groups_appendix} groups regimes according to the similarity of their lagged ENSO responses. This view complements the main lag plot by emphasizing coordinated phase relationships among regimes with related seasonal behavior. The grouped labels in the heatmap are:
\begin{itemize}
    \item W1: clusters 3, 7, 13, 15, 25, 28; W2: cluster 14.
    \item Su1: clusters 4, 11, 12, 16, 18, 19, 23, 24, 27; Su2: clusters 0, 29.
    \item A1: clusters 6, 10, 26; A2: clusters 5, 9, 17, 21.
    \item S1: cluster 8; S2: clusters 1, 20; S3: cluster 22; S4: cluster 2.
\end{itemize}

\begin{center}
    \includegraphics[width=0.95\textwidth]{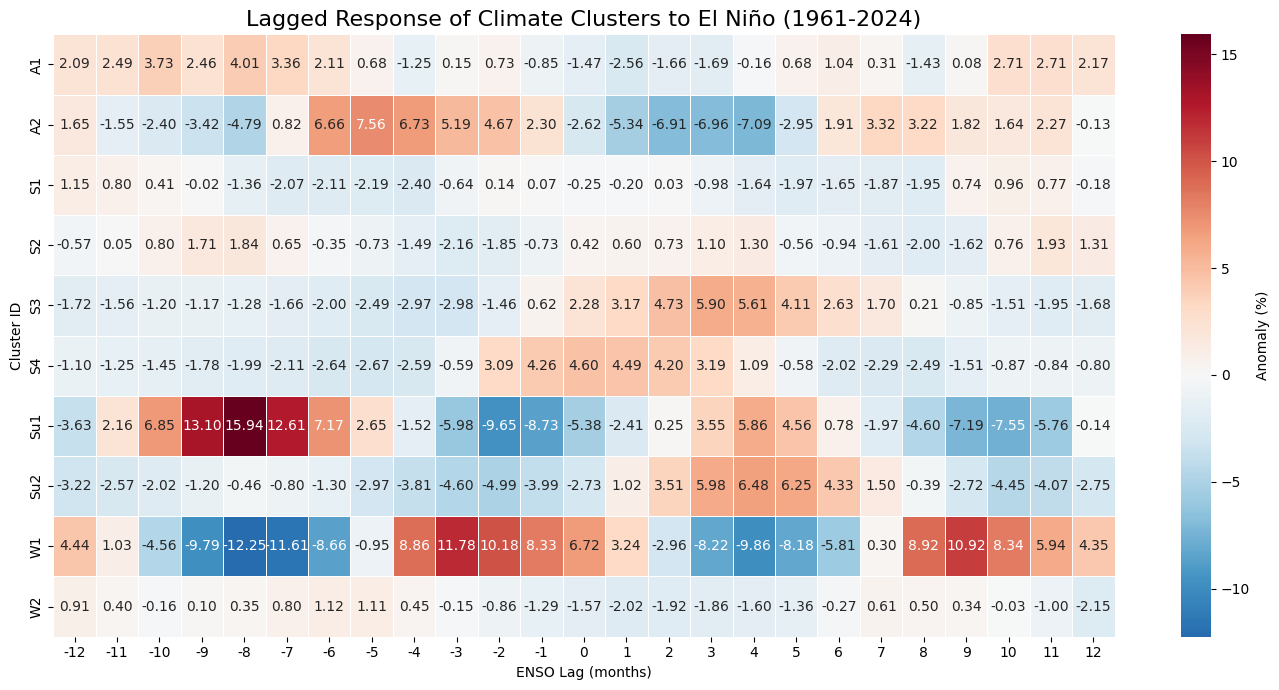}
    \captionof{figure}{Grouped lagged ENSO responses for the learned climate regimes. Regimes are organized according to the similarity of their lagged probability anomalies, highlighting coordinated precursor, synchronous, and delayed behaviors across the ENSO lifecycle.}
    \label{fig:anomalies_lags_groups_appendix}
\end{center}


\end{document}